\pdfoutput=1

\documentclass[11pt]{article}

\usepackage[]{acl}

\usepackage{times}
\usepackage{amsmath}
\usepackage{amssymb}
\usepackage{graphicx}
\usepackage{caption}
\usepackage{subcaption}
\usepackage{xcolor}
\usepackage{float}
\usepackage{booktabs}
\usepackage{natbib}
\usepackage{latexsym}
\usepackage[textsize=scriptsize]{todonotes}
\usepackage{algorithm}
\usepackage{algpseudocode}
\usepackage{algorithmicx}
\usepackage{hyperref}
\usepackage{xr}
\usepackage{bbm}
\usepackage{enumitem}
\usepackage{listings}
\usepackage{multirow}
\usepackage{xcolor}
\usepackage[normalem]{ulem}
\usepackage{spverbatim}
\definecolor{my-blue}{HTML}{1F4C7C}
\definecolor{my-purple}{HTML}{6041A4}
\definecolor{my-orange}{HTML}{E2792E}

\usepackage[T1]{fontenc}

\usepackage[utf8]{inputenc}

\usepackage{microtype}

\title{Improving Cross-task Generalization of Unified Table-to-text Models\\with Compositional Task Configurations}

\author{Jifan Chen$^{1 \ast}$\quad Yuhao Zhang$^2$ \quad Lan Liu$^2$ \quad Rui Dong$^2$\\
\textbf{Xinchi Chen$^2$ \quad Patrick Ng$^2$ \quad William Yang Wang$^2$ \quad Zhiheng Huang$^2$} \\
$^1$The University of Texas at Austin\\
$^2$AWS AI Labs \\
  \texttt{jfchen@cs.utexas.edu}\quad \texttt{\{yhzhang, liuall, ruidong\}@amazon.com}\\ \texttt{\{xcc, patricng, wyw, zhiheng\}@amazon.com} \\}

\begin{document}
\maketitle
\begin{abstract}
There has been great progress in unifying various table-to-text tasks using a single encoder-decoder model trained via multi-task learning~\cite{xie2022unifiedskg}.
However, existing methods typically encode task information with a simple dataset name as a prefix to the encoder.
This not only limits the effectiveness of multi-task learning, but also hinders the model's ability to generalize to new domains or tasks that were not seen during training, which is crucial for real-world applications.
In this paper, we propose \emph{compositional task configurations}, a set of prompts prepended to the encoder to improve cross-task generalization of unified models.
We design the task configurations to explicitly specify the task type, as well as its input and output types.
We show that this not only allows the model to better learn shared knowledge across different tasks at training, but also allows us to control the model by composing new configurations that apply novel input-output combinations in a zero-shot manner.
We demonstrate via experiments over ten table-to-text tasks that our method outperforms the UnifiedSKG baseline by noticeable margins in both in-domain and zero-shot settings, with average improvements of +0.5 and +12.6 from using a T5-large backbone, respectively.
\end{abstract}

\renewcommand{\thefootnote}{\fnsymbol{footnote}}
\footnotetext[1]{Work done during an internship at AWS AI Labs.}
\renewcommand{\thefootnote}{\arabic{footnote}}

\section{Introduction}

\begin{figure}[t]
\centering
\includegraphics[width=0.48\textwidth]{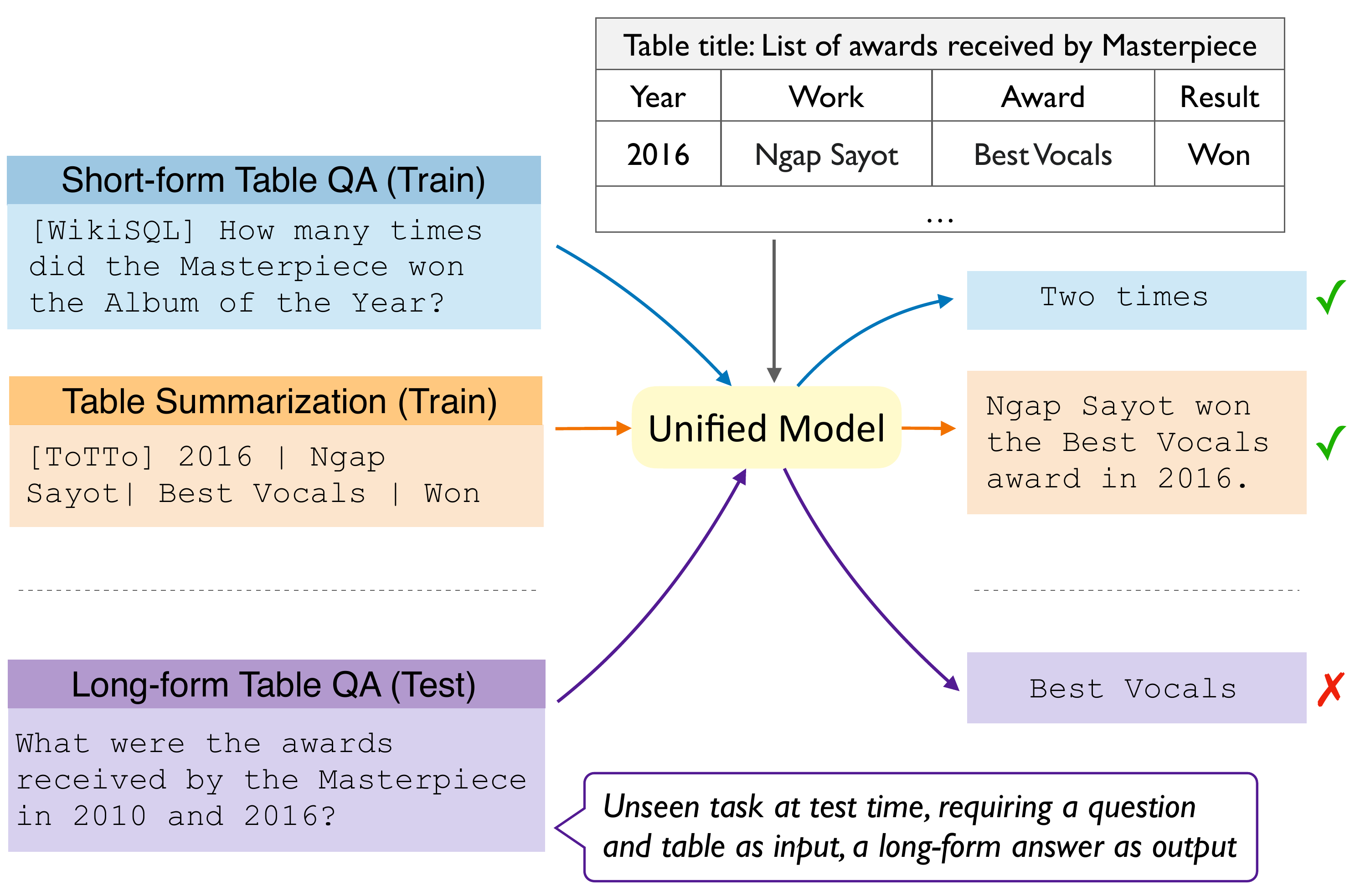}
\caption{
An example of unifying different tasks with a single encoder-decoder model with dataset name as prefix.
The model is trained on short-form table QA and table summarization tasks, and tested on a new long-form table QA task.
As there is a mismatch between the training and test tasks, the model is unable to generalize.
}
    \label{fig:introduction}
\end{figure}

Table-to-text tasks, such as table-based question answering~\cite{pasupat-liang-2015-compositional, herzig-2020-tapas}, summarization~\cite{parikh-etal-2020-totto}, or fact verification~\cite{chen2019tabfact}, are of high interest to the NLP community and have been applied in many real-world applications.
Traditionally, these tasks have been studied individually, with methods commonly optimized for one or a few tasks~\citep{liu2021tapex,shi2021learning}.
However, with the recent popularity of pre-trained transformer models~\citep{raffel2020exploring, lewis-etal-2020-bart, xue-etal-2021-mt5}, there has been a paradigm towards unifying multiple NLP tasks with a single encoder-decoder model~\citep{khashabi-etal-2020-unifiedqa, sanh2022multitask}.
More recently, UnifiedSKG~\citep{xie2022unifiedskg} extended this paradigm to table-to-text tasks by flattening the structured input (e.g., tables) into text format, and unifying all tasks with a T5 model~\citep{raffel2020exploring}.
By training the model over 21 datasets with structured input, it has established new state-of-the-art results for most of these tasks.

Despite the success, existing work often rely on a simple trick to encode task information: the name of the dataset is often used as a prefix to the encoder at both training and test time.
We argue that this overly simplified design has at least two major limitations.
First, since no detailed information about the task is provided, any sharable knowledge between tasks is learned in a latent manner.
Second, with this design, models are trained and evaluated on their abilities to solve specific \emph{datasets}, rather than \emph{tasks}.
As a result, we may see a substantial performance degradation when we apply the model to an unseen task at test time.

Figure~\ref{fig:introduction} illustrates the aforementioned limitations:
a unified model (such as UnifiedSKG) is trained on ~\emph{short-form table QA}~\cite{zhongSeq2SQL2017} and \emph{table-based summarization}~\cite{parikh-etal-2020-totto}, and we want to test the trained model on \emph{long-form table QA}~\cite{nan2022fetaqa}, where the model should take a question and a table as input and output an abstractive sentence as the answer. As there is no way to instruct the model about the information of the new task, the model can only make an educated guess by generating the most plausible text ``Best Vocals'' according to the training datasets, which fails to serve as a good long-form answer. We therefore argue that it is critical to test a table-to-text model's \emph{cross-task generalizability}, which is captured in neither the training methods nor the evaluation setup in existing work.

In this paper, we propose the use of \emph{compositional task configurations}, a set of text prompts prepended to the encoder to improve the cross-task generalizability of unified table-to-text models.
For a given task, we design its configuration prompt to be compositional, describing the task type, input and output types. 
This design offers at least two key advantages.
First, the task configurations explicitly inform the model what is shared between different tasks.
For example, the model is able to learn from the configurations that table-based fact checking and table-based QA share the same inputs but different outputs.
Second and more importantly, using task configurations allows us to have explicit control over the model's behaviors. For the example in Figure~\ref{fig:introduction}, we can now compose a new configuration for long-form table QA at test time to instruct the model to first produce a set of relevant cells and synthesize them to produce a long-form answer, which is within the capabilities of the two training tasks. We discuss this further in the next section.

Our evaluation focuses model's cross-task generalizability.
Specifically, we train our model on 5 table-to-text datasets and test it on an additional set of 5 new datasets that cover either a new domain of an existing task or a new task of which the capabilities can be composed by the ones learned through the 5 training datasets.
We find that our method not only outperforms the strong UnifiedSKG baseline consistently on the 5 in-domain datasets, but also demonstrates much stronger cross-task generalization.
In zero-shot evaluation on the 5 test-only tasks, our model outperforms UnifiedSKG by a substantial margin of +6.5 and +12.6 average scores from using T5-base and T5-large, respectively.
Notably, we find that in zero-shot evaluation on \textsc{FeTaQA}~\citep{nan2022fetaqa}, a long-form table QA task, while the baseline completely fails with a 0.6 F1 score, our method leads to much better generalization, achieving a 21.2 F1 score.
We also show that using the compositional task configurations allows the model to output \emph{supporting table cells} that supplement its final prediction in a zero-shot manner.
Human evaluation of the generated supporting cells for the \textsc{TabFact} dataset reveals that more than 80\% of the generated cells have high relevance to the task.

\begin{figure*}[t]
\centering
\includegraphics[width=\textwidth]{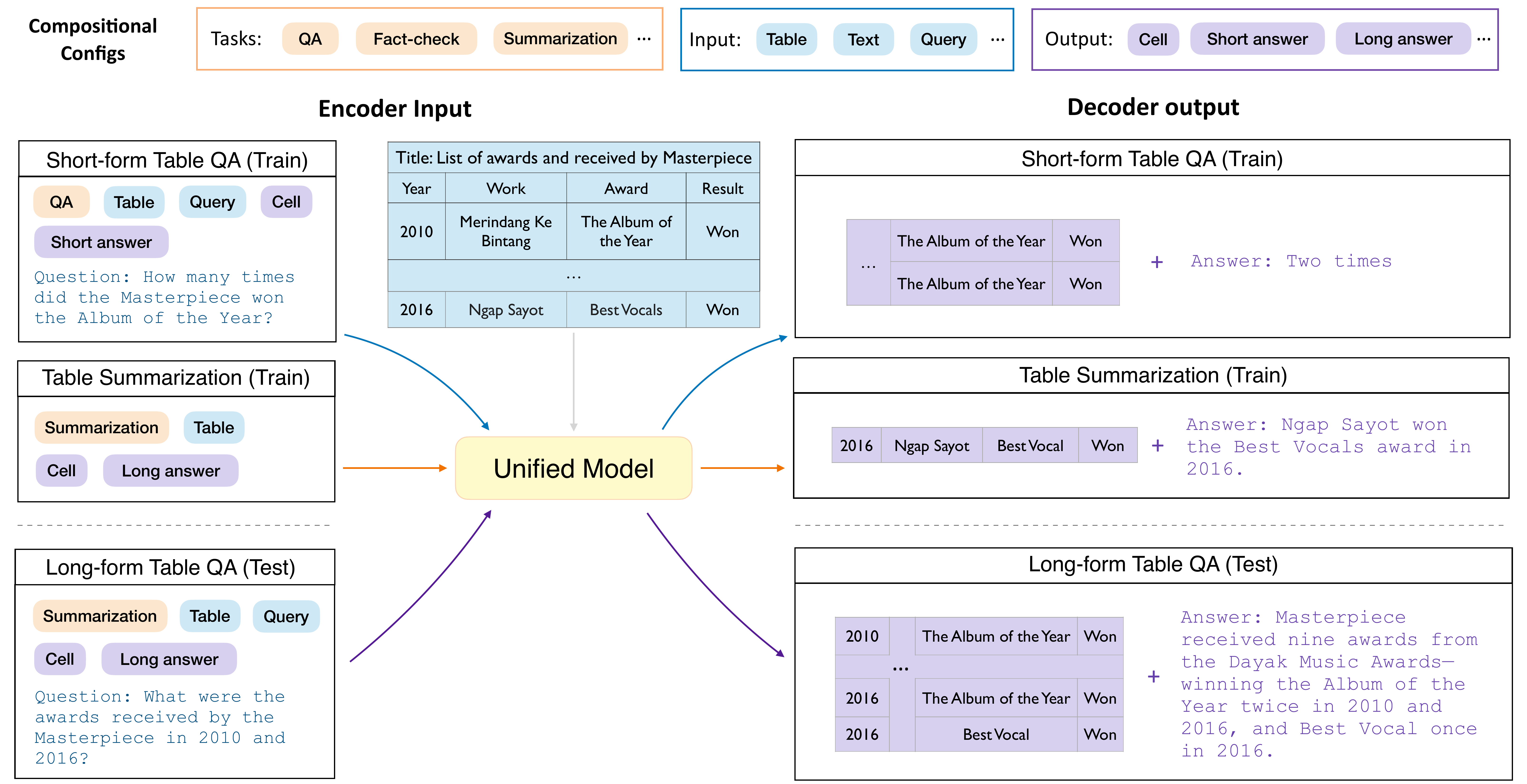}
\caption{Our method demonstrates that we can reformulate the unseen task of long-form table QA as a query-based summarization task by composing a new configuration using the existing ones from the two training tasks. At test time, this composed configuration allows the model to first identify \textcolor{my-purple}{relevant cells} in the table based on the \textcolor{my-blue}{input question} using the cell generation skills learned from \textcolor{my-orange}{short-form table QA}, and then generates a \textcolor{my-purple}{long-form answer} by utilizing the knowledge acquired from \textcolor{my-orange}{table summarization}. For all three tasks, the model also takes the linearized table as input; for simplicity, we omit the linearized table and dataset config in the figure.}
    \label{fig:proposed-work-flow}
\end{figure*}

\section{Method \& Tasks} \label{sec:compositional-task-configs}
Prompting is a natural and feasible way to impose explicit control over the behaviors of pre-trained language models~\cite{wei2021finetuned, chung2022scaling, sanh2022multitask}. In this work, we implement the task configurations as prompts of an encoder-decoder model. Each task configuration contains the following four aspects: task type, input type, output type, and dataset name. The task type is the end goal of a task, e.g., QA and summarization, as shown in Figure~\ref{fig:proposed-work-flow}. Input and output types specify the inputs of the encoder and the outputs of the decoder of table-to-text models, respectively. These types can be compositional, for example, both long-form and short-form table QA in Figure~\ref{fig:proposed-work-flow} require the decoder to output a set of relevant cells and the final answer. The dataset name specifies the dataset used for training. As different datasets can share the same task type, input and output types, we assume that the model is able to learn the shared and the unique knowledge across different datasets by adding the dataset names as configurations. When testing the model on a new dataset, we can simply omit the dataset name since it is not trained.

One of the major advantages of having explicit task configurations is that it enables the model to learn the mapping between a configuration and its behavior. At test time, we can compose a new set of configurations which suits best for an unseen task using the trained configurations. Figure~\ref{fig:proposed-work-flow} demonstrates that by training on short-form table QA, the model learns the ability to generate a set of relevant table cells according to the question and then derive the answer based on those cells. By training on table summarization, the model learns to produce a summary based on a set of table cells.\footnote{Conventionally, when using ToTTo, people feed the cells to the encoder to decode the summary. To make the task more compatible with other table-to-text tasks, we feed the relevant cells as inputs to the decoder and the model is not trained on those cells. } At test time, by reformulating the long-form table QA as a query-based summarization task, our model is able to first generate a set of relevant cells (learned through short-form table QA) and then synthesize those cells to yield a long-form answer (learned from table summarization).

Note that our method is an efficient extension to the original UnifiedSKG model. It only requires a small input prefix, comprising less than 5\% of the total sequence length, making it flexible for generalization to more tasks and datasets.

\begin{table*}[t]
\small
\centering
\begin{tabular}{ l l c  c c c c c c}
\toprule
& Dataset & Task Type &  Input & Output & Unseen? \\
\midrule

\multirow{5}{*}{Train + Test} & \textsc{WikiSQL} & QA (table)  & query + table &	cells + short-form answer & - \\
& \textsc{WikiTQ} & QA (table) & query + table & short-form answer & - \\
& \textsc{SQuAD} & QA (text) & query + passage & short-form answer & - \\
& \textsc{ToTTo} & Summarization &  table & cells + summary & - \\ 
& \textsc{TabFact} & Fact-check  & query + table & binary answer & - \\

\midrule
\multirow{5}{*}{Test-only} & \textsc{NQ-Tables} &  QA (table) & query + table & short-form answer & - \\
& \textsc{HybridQA} & QA (hybrid)  & query + passage + table & short-form answer & \checkmark \\
& \textsc{TAT-QA} & QA (hybrid) & query + passage + table &	short-form answer & \checkmark \\
& \textsc{FeTaQA} & QA (abstractive)  & query + table & cells + long-form answer & \checkmark\\
& \textsc{FEVEROUS} & Fact-check & query + passage + table & binary answer & \checkmark \\
\bottomrule
\end{tabular}
\vspace{-0.2cm}
\caption{Datasets and tasks considered in our experiments.
Tasks that have an input/output combination unseen at training time are marked with \checkmark in the ``Unseen?'' column. 
We include detailed statistics and task configurations applied for each dataset in Appendix~\ref{appendix:dataset-preprocessing} \& \ref{appendix:dataset-task-config} due to space limitation.
} 
\label{tab:dataset-stats}
\end{table*}

\subsection{Datasets and Task Configurations}

A detailed list of our datasets with their task information is shown in Table~\ref{tab:dataset-stats}.
We consider 5 datasets as in-domain datasets for both training and testing: \textsc{WikiSQL}~\citep{zhongSeq2SQL2017}, \textsc{WikiTQ}~\citep{pasupat-liang-2015-compositional}, \textsc{SQuAD}~\citep{rajpurkar-etal-2016-squad}, \textsc{ToTTo}~\citep{parikh-etal-2020-totto} and \textsc{TabFact}~\citep{chen2019tabfact}.
For \textsc{SQuAD} and \textsc{ToTTo}, since no official test set is released, we follow UnifiedSKG~\citep{xie2022unifiedskg} and report results on the official development sets.
In addition, we consider 5 datasets for test only: \textsc{NQ-Tables}~\citep{kwiatkowski-etal-2019-natural, herzig2021open}, \textsc{HybridQA}~\citep{chen-etal-2020-hybridqa}, \textsc{TAT-QA}~\citep{zhu-etal-2021-tat}, \textsc{FeTaQA}~\citep{nan-etal-2022-fetaqa} and \textsc{FEVEROUS}~\citep{aly-etal-2021-fact}.

The test-only evaluation setup aims to assess the effectiveness of our method in enabling the model to generalize to unseen tasks with new compositional configurations, as well as to a new dataset with existing configurations. Specifically, we test if the model can benefit from a combination of input configurations for tasks such as \textsc{HybridQA}, \textsc{TAT-QA}, and \textsc{FEVROUS}, which involve both passages and tables as inputs, despite the model is only trained on one or the other during training. Similarly, we examine if the configuration for \textsc{FeTaQA}, a combination of \textsc{WikiSQL} and \textsc{ToTTo} as shown in Figure~\ref{fig:proposed-work-flow}, allows for explicit control over the model's behaviors, resulting in improved generalization. Finally, we assess the model's ability to generalize to a new dataset, \textsc{NQ-Tables}, which has the same configurations as \textsc{WikiSQL} and \textsc{WikiTQ}.

\begin{figure}[t]
\centering
\includegraphics[width=0.5\textwidth]{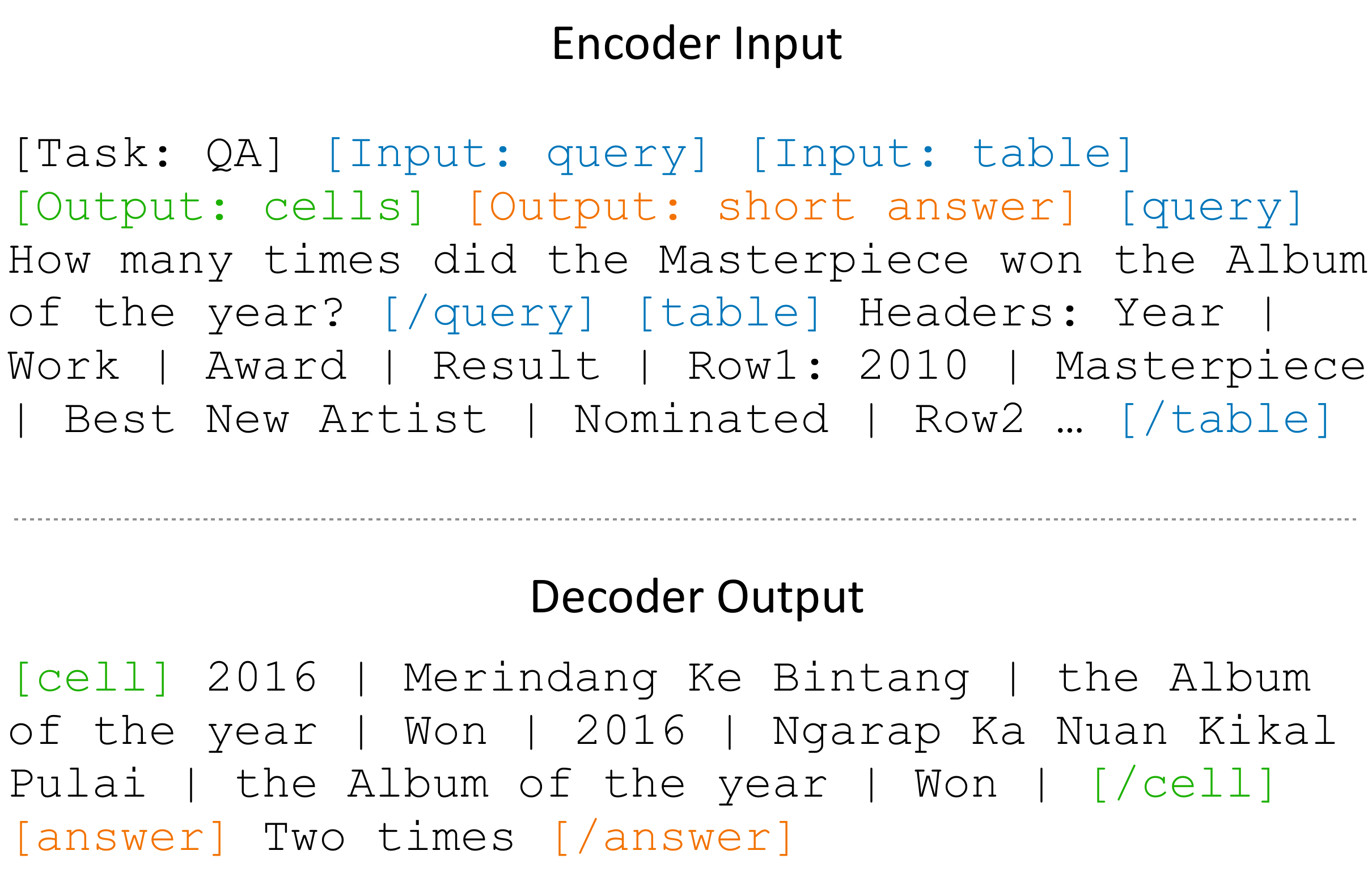}
\caption{An example of the input and output given to the model. After training, the model is able to establish a correspondence between the task configurations and the input/output format.}
\label{fig:input-output-example}
\end{figure}

For all of these datasets, we linearize the tables following the strategy used in UnifiedSKG~\citep{xie2022unifiedskg}. By inserting several special tokens like vertical bars to indicates the boundaries between cells and rows, a table can be linearized as: ``Headers: $h_1 | ... | h_m$, Row 1: $c_{11} | ... | c_{1m}$ ... Row n: $c_{n1} | ... | c_{nm}$''. Here, $h_i$ denotes the $i$th header of a table and $c_{ij}$ denotes the cell content in the $i$th row and the $j$th column. For simplicity, we fix the order of the task configurations to be task type, dataset name, input type, and output type. We prepend the task configuration to the original input of a dataset and feed it to the model. To make our input and output better aligned with the configurations, we also introduce some special markups to separate different parts of inputs and outputs: Figure~\ref{fig:input-output-example} illustrates the actual model's input and output of the short-form table QA example from Figure~\ref{fig:proposed-work-flow}. See Appendix~\ref{appendix:dataset-task-config} for inputs and outputs constructed for all of the datasets and Appendix~\ref{appendix:dataset-preprocessing} for preprocessing details of each dataset.

\section{Experiments}
\paragraph{Experimental settings} We evaluate our method by following the experimental setup shown in Table~\ref{tab:dataset-stats}. We follow the experimental settings of UnifiedSKG~\citep{xie2022unifiedskg} and use T5~\citep{raffel2020exploring} as the backbone of our table-to-text model. To balance the size of different datasets during training, we use the temperature up-sampling method proposed in the original T5 paper and set the temperature to 2. For all experiments, we use a batch size of 128 and AdamW~\citep{loshchilov2018decoupled} as the optimizer with the initial learning rate set to 5e-5. We limit the maximum length of the input, including task configuration and the actual inputs, to be 1024 sentence-piece tokens. 

\begin{table*}[t]

\small
\centering

\begin{tabular}{ c l  c c c c c c}
\toprule
\multicolumn{8}{c}{Zero-shot Test-only Tasks}\\
\midrule
& \multirow{2}{*}{Models} & \textsc{NQ-Tables} & \textsc{FeTaQA} & \textsc{HybridQA} & \textsc{TAT-QA} & \textsc{FEVEROUS} & Avg. \\
& & BLEU & EM & EM & EM & Acc. & -- \\
\midrule
\multirow{3}{*}{T5-base} & Single Task & \it 51.6  & \it 29.9 & \it 54.3 & \it 34.5 & \it 81.3 & \it 50.3 \\
\cmidrule{2-8}
 & UnifiedSKG & 37.8 & 0.6  & 22.5 & 18.2 & 67.5 & 29.3 \\
 & Task Configs (Ours) & \bf 39.4 & \bf 21.0 & \bf 28.9 & \bf 20.8 & \bf 68.9 & \bf 35.8 \\
\midrule
\multirow{3}{*}{T5-large} & Single Task & \it 52.2 & \it 33.0 & \it 56.6 & \it 36.2 & \it 82.1 & \it 52.0 \\
\cmidrule{2-8}
& UnifiedSKG & 42.6 & 0.7 & 34.1 & 20.4 & 41.4 & 27.8 \\
& Task Configs (Ours) & \bf 43.0 & \bf 25.2 & \bf 38.0 & \bf 20.8 & \bf 75.0 & \bf 40.4 \\

\midrule

\multicolumn{8}{c}{In-domain Tasks}\\
\midrule
& \multirow{2}{*}{Models} & \textsc{WikiSQL} & \textsc{WikiTQ} & \textsc{ToTTo} & \textsc{SQuAD} & \textsc{TabFact} & Avg. \\
& & EM & EM & BLEU (dev.) & EM (dev.) & Acc. & -- \\
\midrule
\multirow{3}{*}{T5-base} & Single Task & \it 81.6  & \it 35.8 & \it 36.7 & \it 83.6 & \it 76.1 & \it 62.8 \\
\cmidrule{2-8}
 & UnifiedSKG & 82.9 & 41.1 & 37.2  & 82.5 & 77.1 & 64.2 \\
 & Task Configs (Ours) & \bf 83.5 & \bf 42.5 & \bf 37.4 & \bf 83.0 & \bf 77.5 & \bf 64.8 \\
\midrule
\multirow{3}{*}{T5-large} & Single Task & \it 85.5 & \it 43.4 & \it 37.8 & \it 86.0 & \it 81.0 & 66.7 \\
\cmidrule{2-8}
& UnifiedSKG & 86.0 & 48.5 & \bf 38.7 & 86.1 & 83.0 & 68.5 \\
& Task Configs (Ours) & \bf 86.7 & \bf 50.0 & \bf 38.7 & \bf 86.2 & \bf 83.3 & \bf 69.0 \\

\bottomrule

\end{tabular}

\caption{Zero-shot and in-domain performance of our proposed method (Task Configs) vs. baselines for both T5-base and T5-large. Here, ``EM'' denotes exact match accuracy.
For all tasks we also include the results from single task finetuning as references.
Higher numbers among our method and UnifiedSKG are highlighted in bold.
}
\label{tab:in-domain-plus-zero-shot-results}
\end{table*}

\paragraph{Baseline}
We mainly compare our method against UnifiedSKG~\citep{xie2022unifiedskg}, a strong baseline that was shown to achieve state-of-the-art results on many table-to-text tasks via multi-task training.
In UnifiedSKG, for each task a dataset name is prepended to the encoder during multi-task finetuning as a pseudo task configuration.
For fair comparisons, we re-trained UnifiedSKG models on the five in-domain datasets by using the authors' implementation.%
\footnote{We in fact found that our version of the UnifiedSKG model fine-tuned over the five in-domain datasets outperforms the original authors' version on several datasets, establishing a more competitive baseline. For example, our version achieves 82.9 on WikiSQL and 77.1 on TabFact, whereas the original model by \citet{xie2022unifiedskg} achieves 81.9 and 71.2, respectively.}

\paragraph{Evaluation}
For the in-domain tasks, we simply train on their training sets and evaluate on their test sets.
For the test-only tasks, we evaluate our method in two settings:
1) a \textbf{zero-shot} setting, where we directly apply the model trained on in-domain datasets and use a new set of task configs designed for each test dataset;
2) a \textbf{few-shot} setting, where for each test dataset, we further fine-tune the model using $n$ randomly sampled training examples.
Since we observed that the few-shot training is unstable and heavily depends on the sampled examples, we report average performance from 5 different random seeds (each with a different set of few-shot examples).

\begin{figure*}[t]
\centering
\includegraphics[width=\textwidth]{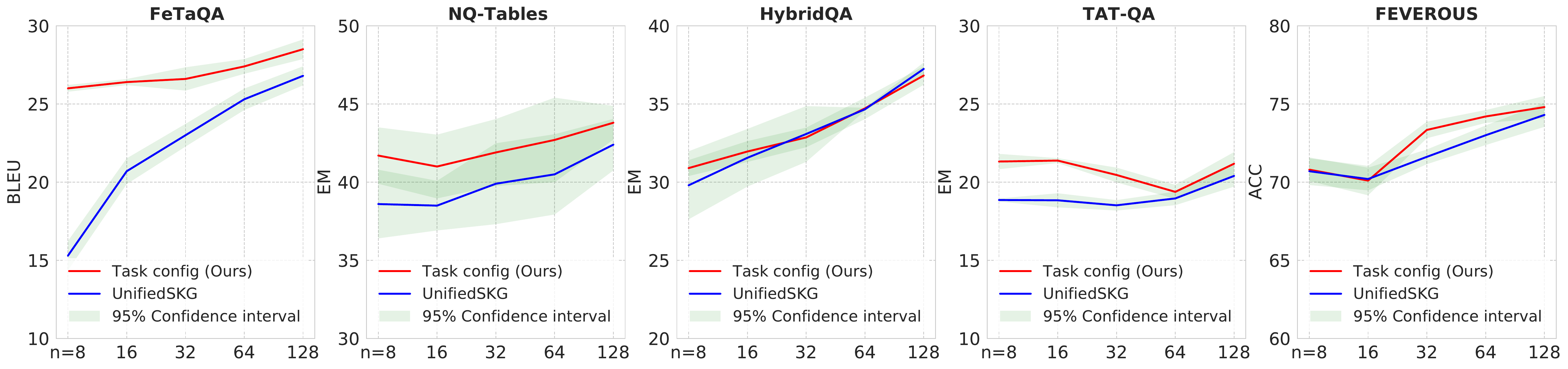}
\caption{Few-shot performance on out-of-domain tasks, with results aggregated from 5 random training runs.
Our method demonstrates better few-shot performance than the baseline in most settings, with the gap reduced as more supervised examples are added.
}
\label{fig:few-shot-performance}
\end{figure*}

\section{Results}
\subsection{Main Results}

We present the in-domain and zero-shot evaluation results for all datasets in Table~\ref{tab:in-domain-plus-zero-shot-results} and the few-shot evaluation results for OOD datasets in Figure~\ref{fig:few-shot-performance}.
We have the following observations:

First, \textbf{using compositional task configs shows much stronger performance on zero-shot datasets unseen at training time} (Table~\ref{tab:in-domain-plus-zero-shot-results}).
For example, the UnifiedSKG baseline fails to generalize at test time to \textsc{FeTaQA}, a long-form table QA task where the input is a question-table pair and the output is a long-form abstractive answer.
This is due to the baseline model having no clue of what format of output should be produced and what knowledge learned through the training datasets should be leveraged for this task.
In contrast, by reformulating the long-form table QA as a query-based summarization task and composing the input configurations to be \emph{table} and \emph{query} as well as the output task configs to be \emph{relevant cell} and \emph{summary}, our method notably improves the zero-shot performance and closes the gap between zero-shot and single-task finetuning results.
Note that among the zero-shot datasets, \textsc{NQ-Tables} represents a new dataset for an existing task (short-form table QA), whereas others represent new tasks unseen at training.
Nevertheless, we found the improvements to be consistent for all zero-shot datasets, with average improvements of +6.5 and +12.6 for base and large models, respectively.

Second, in most cases, \textbf{using compositional task configs consistently improves the in-domain performance} over the UnifiedSKG baseline and single-task training (Table~\ref{tab:in-domain-plus-zero-shot-results}).
The observation is consistent for both base and large model sizes, with average improvements of +0.6 and +0.5 over UnifiedSKG, respectively.
The improvement over single-task fine-tuning is even greater for all datasets.
One explanation to this improvement is that by adding task configs we explicitly encourage the model to learn the shared knowledge between different tasks and datasets.

Last but not the least, in few-shot evaluation (as depicted in Figure~\ref{fig:few-shot-performance}), we find that \textbf{using task configurations has improved few-shot learning performance for most test-time tasks}.
Overall the difference between our method and the UnifiedSKG baseline is particularly notable when the number of supervised examples ($n$) is small; and the performance gap diminishes for \textsc{HybridQA}, \textsc{TAT-QA}, and \textsc{FEVEROUS} as $n$ gets larger. One possible explanation is that the prior captured by the task configurations during training is not closely aligned with these three datasets, when $n$ getting larger, the prior introduced by the task configurations is gradually overridden by knowledge learned from the supervised data.

\subsection{Ablation of Task Configs at Training time} The impact of individual configurations on model performance was evaluated by removing one configuration at a time during training. The results, presented in Table~\ref{tab:config-ablation-training}, indicate that the removal of the \textbf{output type} resulted in the largest performance drop, as the model was only able to guess the desired output type based on learned parameters. The removal of the \textbf{input type} had the least impact on performance. This is likely due to the fact that learning the representation of the two input types was not difficult for the model, and explicitly informing the model about the input type does not provide significant benefit, as observed in the previous section. The removal of the \textbf{dataset type} also results in a performance drop, particularly on the \textsc{NQ-Tables} dataset, indicating that even when the task type, input, and output are the same, including the dataset type helps the model learn dataset-independent knowledge more effectively. The removal of the \textbf{task type} results in a complete failure on the \textsc{FeTaQA} dataset, demonstrating that in some cases, all configurations are necessary to produce the correct form of output. A more detailed discussion of these findings can be found in section~\ref{sec:limitations}.

\label{appendix:alation-training-time}

\begin{table}[t]
\small
\centering
\setlength{\tabcolsep}{2.5pt}
\begin{tabular}{ l c c c c c c }
\toprule

Models & \textsc{FeTaQA} & \textsc{NQ} & \textsc{Hybrid} & \textsc{TAT} & \textsc{FEVR} & Avg. \\
\midrule
Full configs & 21.0 &	39.4 & 28.9 & 20.8 & 68.9 & \bf 35.8 \\
- dataset  & 21.2 & 36.1 & 25.3 & 19.8 & 68.1 & 34.1 \\
- task type & 0.4 & 38.4 & 28.3 & 21.3 & 68.5 & 31.4 \\ 
- input & 20.3 & 39.5 & 29.1 & 19.4 & 68.3 & 35.3 \\
- output & 17.3 & 34.8 & 17.9 & 16.1 & 68.2 & 30.9 \\
\bottomrule
\end{tabular}
\caption{Ablation of task configurations during training. We only report zero-shot performance here. We see removing either one of the configurations cause a performance drop.
}
\label{tab:config-ablation-training}
\end{table}

\subsection{Ablation of Task Configs at Test time}
\begin{figure}[t]
\centering
\includegraphics[width=0.45\textwidth]{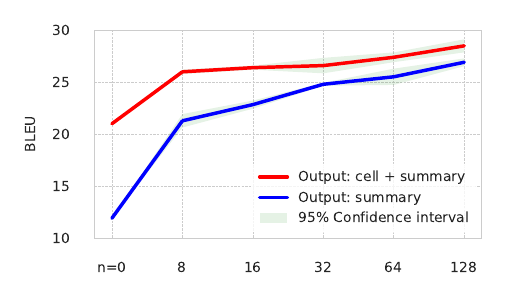}
\caption{Ablation of the output configurations on \textsc{FeTaQA}. The model achieve better cross-task performance by generating a set of relevant cells in both zero-shot and few-shot settings. Here, ``n=0'' denotes zero-shot setting.}
    \label{fig:ablation-output-type}
\end{figure}

While our method demonstrates much stronger zero-shot task performance, it is crucial to understand the extent to which input and output configurations contribute to this success, particularly for tasks involving hybrid input or output types that are not present during training.
To examine the contributions of input configurations, we remove each configuration from the hybrid tasks (\textsc{HybridQA}, \textsc{TAT-QA} and \textsc{FEVEROUS}) at test time, with results shown in Table~\ref{tab:ablation-input-type}.
We found that deleting either of the input configurations results in a performance drop in most cases, and the drop is quite notable when the table and passage input configurations are removed together.
This suggests that the input configuration captures useful priors about the input during training, and \textbf{different configurations can be combined to yield better performance in the zero-shot transfer to hybrid tasks.}
We also observe a similar trend in Figure~\ref{fig:ablation-output-type} where we test the model performance by removing the \emph{cell} output configurations for \textsc{FeTaQA} (thereby skipping cell generation). 
We see that in both zero-shot and few-shot settings, model performance drops by a large margin.
This shows not only that the model can generate different outputs by combining the output configurations, but also that it can better utilize the prior captured by the configurations to improve task performance.

\subsection{Human Evaluation of Generated Cells}

\begin{figure*}[t]
\centering
\includegraphics[width=\textwidth]{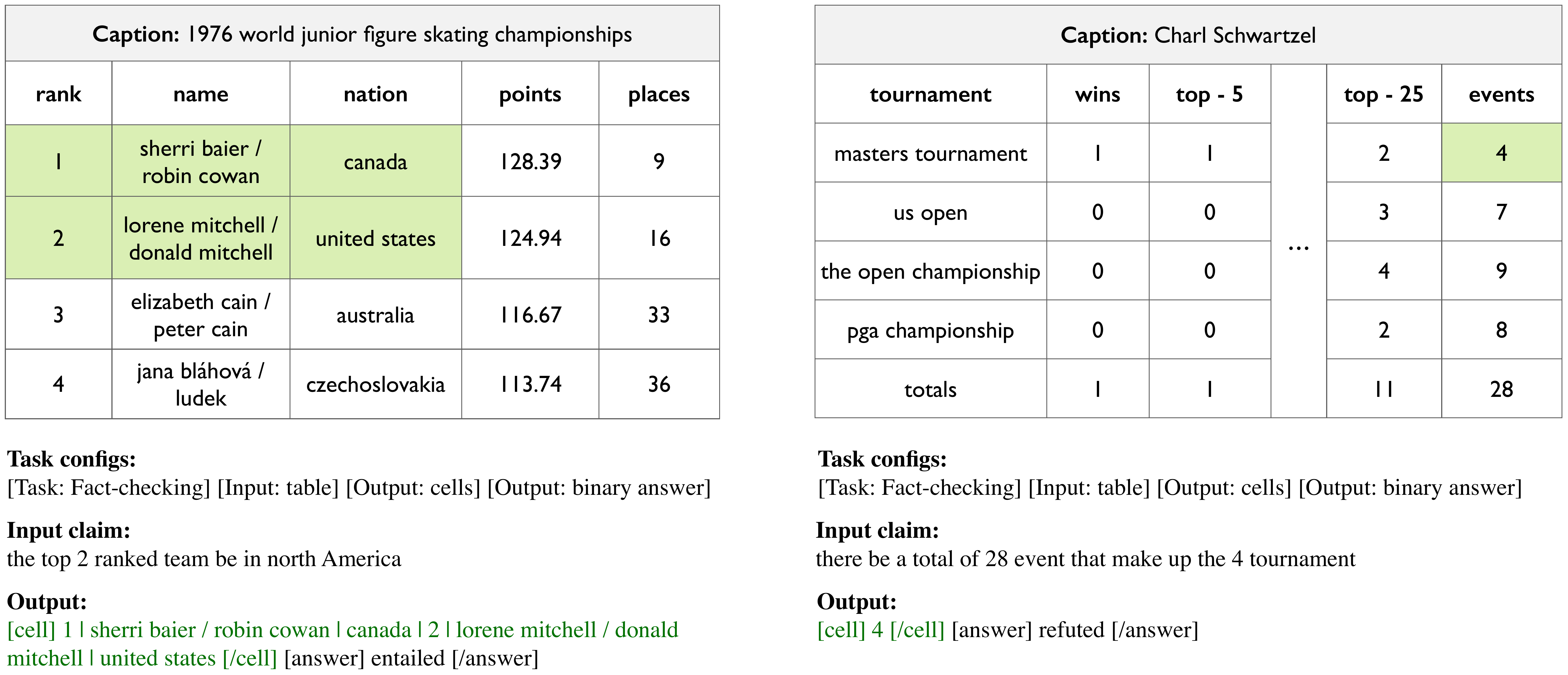}
\caption{
Two examples from the development set of \textsc{TabFact} where we force the model to produce relevant cells by changing the output configurations.
For the left example, the model produces a set of complete and relevant cells that help understand its ``entailed'' prediction.
For the right example, the model is misled by the number ``4'' when generating the cells, and we are able to check that the final answer ``refuted'' is wrong.
}
\label{fig:tabfact-case-study}
\end{figure*}

In addition to the strong task generalizability, a key advantage of applying the proposed task configurations to table-to-text tasks is that we can modify the task configurations to output more results for improved explainability, even when such a configuration combination is never seen at training time.
An example of this is for the table-based fact verification task, \textsc{TabFact}, instead of only generating a binary label, we can extend the output configuration to include a \emph{cell} component that can serve as supporting evidence of the binary prediction. We include two examples of this setting in Figure~\ref{fig:tabfact-case-study}.

To understand how well our model can generate supporting cells without ever being trained for it, we conduct a human study over 50 randomly sampled outputs from the \textsc{TabFact} dataset.
We ask human annotators to manually evaluate the generated cells based on their level of \textbf{relevance} and \textbf{completeness}.
Relevance denotes the usefulness of the generated cells in verifying a claim (precision) and completeness refers to the extent to which all of the relevant cells are generated (recall). 
For each aspect, we ask the annotators to select between three labels that characterize its degree: ``full'', ``partial'' or ``none''.
Three of the authors conduct the annotations, achieving 0.72 and 0.80 Fleiss Kappa~\cite{fleiss1971measuring} for relevance and completeness, respectively. We conduct majority vote to get the consensus label and the results are shown in Table~\ref{tab:cell-generation-eval}.
Overall we found that the model is able to generate cells with high relevance (with 72\% examples being fully relevant generations), but struggle with full completeness (with 34\% fully complete).

\begin{table}[t]
\small
\centering
\setlength{\tabcolsep}{3pt}
\begin{tabular}{ l c c c c}
\toprule
Models & \textsc{HybridQA} & \textsc{TAT-QA} & \textsc{FEVR} & Avg.\\
\midrule
Full Configs & 28.9 & \bf 20.8 & \bf 68.9 & \bf 39.5 \\
- input:passage  & 28.8 & 19.8 & 68.6 & 39.1 \\ 
- input:table  & \bf 29.4 & 20.3 & 66.7 & 38.8 \\
- input:all  &	27.3 & 19.3 & 66.1 & 37.6 \\
\bottomrule
\end{tabular}
\caption{Ablation of the input configurations at test time. The inputs of the three datasets include both table and passage. We show the model can achieve better performance by combining two input configurations.}
\label{tab:ablation-input-type}
\end{table}

\begin{table}[t]
\small
\centering
\begin{tabular}{ l c c c }
\toprule

            & Full & Partial & None  \\
\midrule
Relevance & 72\% & 10\% & 18\%  \\
Completeness  & 34\% & 48\% & 18\%  \\

\bottomrule
\end{tabular}
\caption{Human evaluation results of the zero-shot cell generation quality for the \textsc{TabFact} task.}
\label{tab:cell-generation-eval}
\end{table}

\section{Related Work}
\paragraph{Table-to-text tasks} Table-based tasks, including table-based question answering~\citep{pasupat-liang-2015-compositional, zhongSeq2SQL2017, chen-etal-2020-hybridqa, cheng-etal-2022-hitab, zhao-etal-2022-multihiertt}, table-based fact-checking~\citep{chen2019tabfact, aly-etal-2021-fact}, table summarization~\citep{parikh-etal-2020-totto, suadaa-etal-2021-towards, moosavi2021learning}, have gained increasing attention in recent years. A flurry of work using transformer-based structure explored modeling table structure via pretraining, for example, TabTransformer~\citep{huang2020tabtransformer}, VIME~\citep{yoon2020vime}, TABBIE~\citep{iida-etal-2021-tabbie}, TaBERT~\citep{yin-etal-2020-tabert}, TUTA~\citep{wang2021tuta}, TabT5~\citep{andrejczuk2022table}, and TableFormer~\citep{yang-etal-2022-tableformer}. 

 Our work mainly focuses on table-to-text tasks but the proposed framework is capable of generalizing to a broader range of tasks and datasets.

\paragraph{Task unification} There have been a vein of work that tries to solve various NLP tasks using a single model. This includes encoder-decoder models like T5~\citep{raffel2020exploring}, UnifiedQA~\citep{khashabi-etal-2020-unifiedqa}, UnifiedQA2~\citep{khashabi2022unifiedqa}, UnifiedSKG~\citep{xie2022unifiedskg}; decoder-only models driven by prompts, for example, GPT3~\citep{NEURIPS2020_1457c0d6}, Codex~\citep{chen2021evaluating}, PaLM~\citep{chowdhery2022palm}.
Our work extends UnifiedSKG by using an encoder-decoder model as the backbone and designing prompts to encourage better knowledge sharing between different tasks and enable control over the model's behaviors.

\paragraph{Cross-task generalization with pretrained models} Various efforts have been made to improve the ability of unified models to generalize to new tasks and datasets, including fine-tuning using a wide range of natural language instructions~\citep{chung2022scaling, sanh2022multitask, wei2021finetuned, zhong-etal-2021-adapting-language}, better design of prompts in zero-shot and few-shot setting~\citep{wei2022chain, zhou2022least, kojima2022large}. Our proposed method is most relevant to Macaw~\citep{tafjord2021general}, ProQA~\citep{zhong-etal-2022-proqa}, and SchemaPro~\citep{zhong2022improving}, which also utilize explicit task descriptions to facilitate knowledge sharing between various NLP tasks. Our work differs in two main aspects: 
(1) Our work focuses on compositional generalization at test time, examining whether the model can combine different configurations from multiple tasks during training to generalize to unseen tasks at test time. (2) Our work focuses on table-to-text tasks. 

\section{Limitations} \label{sec:limitations}

\paragraph{Task configurations are entangled with the full model parameters.} In our ablation study of task configurations at training time (Table~\ref{tab:config-ablation-training}), we see that when training without \textbf{task type}, the model fails to generalize to \textsc{FeTaQA}. Upon examining the model output, we find that although we change the output configuration to ``long answer'', the model still produces a short-form answer. This indicates that model behaviors are not always aligned with a single configuration, leading us to question the extent to which each individual configuration influences the model. In order to have better and more interpretable control over the models, one potential avenue for future research is to develop pluggable task configurations, where each configuration controls a more atomic function of the model and can be plugged, unplugged, and combined to yield different model behaviors.

\paragraph{Our exploration scope is limited to table-to-text tasks.} Due to the constraints of the computational resources, we haven't explored joint training with a broader range of other NLP tasks. We think with some modifications, such as the inclusion of dataset domains in the configuration set, it would be possible to extend our approach to additional datasets and tasks.

\section{Conclusion} \label{sec:conclusion}
We introduced compositional task configurations for unified table-to-text models. Compared to existing unified encoder-decoder models that simply use dataset names as input prefix, compositional task configurations allow us to specify the task type, input, and output types at a finer level, which improve multi-task learning effectiveness and cross-task generalization. Further, we showed that our method allows fine-grained control over the model's generation, improving explainability via generating high-quality supporting table cells.

\bibliography{anthology,custom}
\bibliographystyle{acl_natbib}

\appendix

\section{Dataset Statistics and Preprocessing}
\label{appendix:dataset-preprocessing}

\begin{table}[t]
\small
\centering
\begin{tabular}{ l l c c c }
\toprule

& Dataset & Train & Dev & Test  \\
\midrule
 \multirow{5}{*}{Train+Test} & \textsc{WikiSQL} & 56,355 & 8,421 & 15,878 \\
 & \textsc{WikiTQ}  & 11,321 & 2,810 & 4,344 \\
 & \textsc{SQuAD}  & 87,599 & 10,570 & -- \\
 & \textsc{ToTTo}  & 120,761 & 7,700 & -- \\ 
 & \textsc{TabFact} & 92,283 & 12,792 & 9,750 \\

\midrule
 \multirow{5}{*}{Test-only} & \textsc{NQ-Tables} & -- & -- & 549 \\
 & \textsc{HybridQA} & -- & 3,466 & -- \\
  & \textsc{TAT-QA} & -- & -- & 1,669 \\
 & \textsc{FeTaQA} & -- & -- & 2,003 \\
 & \textsc{FEVEROUS} & -- & 7,890 & -- \\
\bottomrule
\end{tabular}
\caption{Dataset statistics of the datasets we used in the paper. Note that for the test-only datasets, except for few-shot experiments, only the test splits of the original datasets are used. For \textsc{SQuAD}, \textsc{HybridQA}, \textsc{FEVEROUS}, and \textsc{ToTTo}, as no public test set is offered, we evaluate the model on the original development sets following UnifiedSKG~\citep{xie2022unifiedskg}.
For \textsc{NQ-Tables}, we use a modified version of it in our experiments as described in Appendix~\ref{appendix:dataset-preprocessing}.
}
\label{tab:dataset-statistics}
\end{table}

The statistics of the datasets we used in this paper are shown in Table~\ref{tab:dataset-statistics}.
To fit the data into the encoder, for all datasets, we limit the max length of each cell to be 15 sentence-piece tokens.
If the length (measured by sentence-piece tokens) of the linearized table is longer than 1024, we truncate random rows to reduce the table size.

The original annotation of \textsc{WikiSQL} dataset does not include the relevant cells.
We extract the relevant cells by executing the accompanied SQL query annotations.
In most cases, the relevant cells equal to the final answer annotations; for the rest of the cases, aggregations or numerical operations need to be run to obtain the final answer.
During training, we also create another version of the \textsc{WikiSQL} dataset, in which we exclude the relevant cells and only use the final answer as supervision to improve output diversity.
We use both versions at training time.

\textsc{NQ-Tables} is a table-based QA dataset derived from the NaturalQuestions dataset \citep{kwiatkowski-etal-2019-natural} and was originally released by \citet{herzig2021open}.
The original test set of \textsc{NQ-Tables} contains 966 unique examples.
In our experiments, to make the dataset more compatible with other table-based QA tasks, we evaluate on a customized version of \textsc{NQ-Tables} where we only include an example if the answer is uniquely locatable as one or more table cells.
This filtering step results in 549 unique triples of table, question and answers.

To make \textsc{ToTTo} more compatible with other table-to-text tasks, we feed the selected cells as inputs to the decoder, as we mentioned in Section~\ref{sec:compositional-task-configs}.
Also, we find it helpful to create a reversed version of the \textsc{ToTTo} dataset, where we treat the annotated summary and the table as input and let the model predict the relevant cells.
We add both versions of \textsc{ToTTo} to the training of all models, including the baseline.

\section{Task Configurations Applied for Each Dataset}
\label{appendix:dataset-task-config}

Below we list the task configurations applied to all datasets.
For each dataset, we present input to the encoder and output from the decoder separately.
For encoder, we include the template of the full input, including task configurations as well as how the dataset input is structured (with actual data replaced by ``...'').
For decoder, we include how we structure the annotated output during training and how we parse the output during testing.

\subsection{WikiSQL}
\textbf{Encoder:}
\begin{lstlisting}
[Task: QA] [Dataset: WikiSQL] [Input: query] [Input: table] [Output: cells] [Output: short answer] [query] ... [/query] [table] ... [/table]
\end{lstlisting}
\textbf{Decoder:}
\begin{lstlisting}
[cell] ... [/cell] [answer] ... [/answer]
\end{lstlisting}

\subsection{WikiTQ}
\textbf{Encoder:}
\begin{lstlisting}
[Task: QA] [Dataset: WikiTQ] [Input: query] [Input: table] [Output: short answer] [query] ... [/query] [table] ... [/table]
\end{lstlisting}
\textbf{Decoder:}
\begin{lstlisting}
[answer] ... [/answer]
\end{lstlisting}

\subsection{SQuAD}
\textbf{Encoder:}
\begin{lstlisting}
[Task: QA] [Dataset: SQuAD] [Input: query] [Input: passage] [Output: short answer] [query] ... [/query] [passage] ... [/passage]
\end{lstlisting}
\textbf{Decoder:}
\begin{lstlisting}
[answer] ... [/answer]
\end{lstlisting}

\subsection{ToTTo}
\textbf{Encoder:}
\begin{lstlisting}
[Task: Summarization] [Dataset: ToTTo] [Output: cells] [Output: long answer]
\end{lstlisting}
\textbf{Decoder:}
\begin{lstlisting}
[cell] ... [/cell] [answer] ... [/answer]
\end{lstlisting}

 \subsection{TabFact}
 \textbf{Encoder:}
\begin{lstlisting}
[Task: Fact-checking] [Dataset: TabFact] [Input: query] [Input: table] [Output: binary answer] [query] ... [/query] [table] ... [/table] 
\end{lstlisting}
\textbf{Decoder:}
\begin{lstlisting}
[answer] ... [/answer] 
\end{lstlisting}

 \subsection{NQ-Tables}
 \textbf{Encoder:}
\begin{lstlisting}
[Task: QA] [Input: query] [Input: table] [Output: short answer] [query] ... [/query] [table] ... [/table]
\end{lstlisting}
\textbf{Decoder:}
\begin{lstlisting}
[answer] ... [/answer] 
\end{lstlisting}

 \subsection{HybridQA}
 \textbf{Encoder:}
\begin{lstlisting}
[Task: QA] [Input: query] [Input: table] [Input: passage] [Output: short answer] [query] ... [/query] [table] ... [/table] [passage] ... [/passage]
\end{lstlisting}
\textbf{Decoder:}
\begin{lstlisting}
[answer] ... [/answer] 
\end{lstlisting}

 \subsection{TAT-QA}
 \textbf{Encoder:}
\begin{lstlisting}
[Task: QA] [Input: query] [Input: table] [Input: passage] [Output: short answer] [query] ... [/query] [table] ... [/table] [passage] ... [/passage]
\end{lstlisting}
\textbf{Decoder:}
\begin{lstlisting}
[answer] ... [/answer] 
\end{lstlisting}

 \subsection{FeTaQA}
 \textbf{Encoder:}
\begin{lstlisting}
[Task: Summarization] [Input: query] [Input: table] [Output: cells] [Output: long answer] [query] ... [/query] [table] ... [/table] 
\end{lstlisting}
\textbf{Decoder:}
\begin{lstlisting}
[cell] ... [/cell] [answer] ... [/answer] 
\end{lstlisting}

 \subsection{FEVEROUS}
 \textbf{Encoder:}
\begin{lstlisting}
[Task: Fact-checking] [Input: query] [Input: table] [Input: passage] [Output: binary answer] [query] ... [/query] [table] ... [/table] [passage] ... [/passage]
\end{lstlisting}
\textbf{Decoder:}
\begin{lstlisting}
[answer] ... [/answer]
\end{lstlisting}

\subsection{WikiSQL-Answer-only}
\textbf{Encoder:}
\begin{lstlisting}
[Task: QA] [Dataset: WikiSQL] [Input: query] [Input: table] [Output: short answer] [query] ... [/query] [table] ... [/table]
\end{lstlisting}
\textbf{Decoder:}
\begin{lstlisting}
[answer] ... [/answer]
\end{lstlisting}

\subsection{ToTTo-Reverse}
 \textbf{Encoder:}
\begin{lstlisting}
[Task: Cell-generation] [Input: query] [Input: table] [Output: cell] [query] ... [/query] [table] ... [/table]
\end{lstlisting}
\textbf{Decoder:}
\begin{lstlisting}
[cell] ... [/cell]
\end{lstlisting}

\end{document}